\documentclass[11pt]{article}
\usepackage{algorithm}

\usepackage[final]{acl}
\usepackage{enumitem}
 \usepackage{amsmath} 
\usepackage{times}
\usepackage{latexsym}
\usepackage{multirow}
\usepackage[T1]{fontenc}
\usepackage{booktabs}
\usepackage[utf8]{inputenc}

\usepackage{microtype}

\usepackage{inconsolata}

\usepackage{graphicx}

\title{A Hybrid Supervised-LLM Pipeline for Actionable Suggestion Mining in Unstructured Customer Reviews}

\author{
Aakash Trivedi$^{1}$ \quad
Aniket Upadhyay$^{1}$ \quad
Pratik Narang$^{1}$ \quad
Dhruv Kumar$^{1}$ \quad
Praveen Kumar$^{2}$ \\
$^{1}$Department of Computer Science \& Information Systems, \\
Birla Institute of Technology and Science, Pilani, India \\
\texttt{f20191076P@alumni.bits-pilani.ac.in,} \\
\texttt{\{p20241007, pratik.narang, dhruv.kumar\}@pilani.bits-pilani.ac.in,} \\
$^{2}$Birdeye Inc., Palo Alto, California, USA \\
\texttt{praveen.kumar1@birdeye.com}
}

\begin{document}
\maketitle

\begin{abstract}
Extracting actionable suggestions from customer reviews is essential for operational decision-making, yet these directives are often embedded within mixed-intent, unstructured text. Existing approaches either classify suggestion-bearing sentences or generate high-level summaries, but rarely isolate the precise improvement instructions businesses need.
We evaluate a hybrid pipeline combining a high-recall RoBERTa classifier trained with a precision–recall surrogate to reduce unrecoverable false negatives with a controlled, instruction-tuned LLM for suggestion extraction, categorization, clustering, and summarization. Across real-world hospitality and food datasets, the hybrid system outperforms prompt-only, rule-based, and classifier-only baselines in extraction accuracy and cluster coherence. Human evaluations further confirm that the resulting suggestions and summaries are clear, faithful, and interpretable.
Overall, our results show that hybrid reasoning architectures achieve meaningful improvements fine-grained actionable suggestion mining while highlighting challenges in domain adaptation and efficient local deployment.


\end{abstract}

\section{Introduction}

Customer reviews contain valuable signals for service improvement, but explicit suggestions,
concrete requests for what should be fixed, added, or improved are typically rare and embedded
within long, mixed-intent narratives. In this work, we define an actionable suggestion as an explicit, business-directed suggestion that specifies a concrete operational change (e.g., “Add more vegetarian options”), rather than general opinions, complaints, or advice to other customers. Automatically identifying these actionable spans remains
challenging, reviews blend praise, complaints, stories, and user-to-user advice, making heuristic or
manual approaches unreliable at scale.

Prior work on suggestion mining has focused largely on sentence-level detection
\citep{negi2015, wicaksono2013, dong2017attention}, which identifies the presence of a suggestion but
does not extract the actionable phrase, handle multi-sentence directives, or distinguish
business-directed improvements from general opinions. Transformer-based and domain-adaptive models
\citep{joshi2020spanbert, riaz2024} improve detection but still frame the task as classification rather
than full extraction.

Related research in opinion summarization and theme modeling
\citep{angelidis2021gSum, mukku2024, nayeem2024} captures high-level topics, but does not surface the
specific improvements needed for operational decision-making. Meanwhile, LLMs offer strong structured extraction
capabilities \citep{ouyang2022training}, yet LLM-only methods suffer from
hallucination \citep{ji2023survey}, inconsistent span boundaries \citep{koto2022span}, and low recall
for infrequent suggestion types. Conversely, rule-based or classifier-only systems are brittle and
lack generalization.

We investigate whether a hybrid architecture, pairing a high-recall supervised classifier with
controlled LLM-based extraction, categorization, clustering, and summarization can more reliably
surface actionable suggestions from reviews. We frame this as end-to-end \emph{actionability
extraction}, detecting suggestion-bearing reviews, isolating explicit improvement directives, grouping
them semantically, and producing concise summaries suitable for decision-making.

Our contributions are:
\begin{itemize}
    \item A recall-oriented RoBERTa classifier trained with a precision–recall surrogate objective to
    reduce unrecoverable false negatives, while maintaining comparable precision.
    \item An instruction-tuned, quantized LLM for controlled extraction, categorization, clustering,
    and summarization.
    \item Extensive comparisons against prompt-only LLMs, rule-based systems, classifier-only
    pipelines, and end-to-end LLM methods.
    \item Comprehensive evaluation of extraction, category assignment, clustering, and summarization
    using automatic metrics, human judgments, and ablations.
\end{itemize}

By focusing on explicit, operationally meaningful suggestions rather than generic opinions, we show
that a hybrid approach mitigates the weaknesses of classifier-only and LLM-only systems, offering an approach suitable for large-scale operational settings.

\section{Related Work}

\subsection{Suggestion Mining}
Early work framed suggestion mining as binary classification, using benchmarks by \citet{negi2015} and linguistic-pattern methods \citep{wicaksono2013}. Neural models with attention \citep{dong2017attention} and transformer variants such as TransLSTM \citep{riaz2024} improved detection, while span-based architectures (e.g., SpanBERT; \citealp{joshi2020spanbert}) support finer extraction. However, these systems largely detect suggestion presence rather than extracting explicit actionable spans or handling multi-sentence suggestions.

\subsection{Opinion Summarization and Theme Modeling}
Opinion summarization condenses reviews into themes or aspect-level insights. Topic models \citep{blei2003latent, dieng2020topic} and modern abstractive systems \citep{brazinskas2020learning, angelidis2021gSum} produce high-level representations, and domain-specific models such as InsightNet \citep{mukku2024} and LFOSum \citep{nayeem2024} cluster user opinions. Yet these approaches emphasize broad aspects rather than the precise improvements customers request, limiting actionability.

\subsection{Hybrid Approaches and Multi-Stage Reasoning}
Hybrid pipelines combining targeted classifiers with downstream reasoning are common in fact verification \citep{thorne2018fever}, relation extraction \citep{zhou2018joint}, and retrieval-augmented QA \citep{chen2017reading}. Surveys highlight classifier-driven constraints as a method to reduce LLM hallucination \citep{wu2023survey}. However, such hybridization has not been explored for actionable suggestion extraction nor evaluated across downstream stages (clustering, summarization).

\subsection{LLMs for Structured Extraction}
Instruction-tuned LLMs, including GPT models \citep{ouyang2022training}, LLaMA \citep{touvron2023llama}, Mistral \citep{jiang2023mistral}, and Gemma \citep{gemma2024model} enable strong structured extraction, yet LLM-only pipelines remain prone to hallucination \citep{ji2023survey}, unstable span boundaries \citep{koto2022span}, and degraded performance on large input batches. Our approach mitigates these issues using classifier gating and tightly controlled prompting in a multi-stage pipeline.

\subsection{Positioning}
Where prior work targets suggestion detection, theme discovery, or high-level summarization, we focus on extracting \emph{explicit, actionable} suggestions and organizing them into interpretable structures. Our evaluation spans classification, extraction, categorization, clustering, summarization, cross-domain generalization, and ablations, providing the comprehensive study of end-to-end actionability extraction.

\section{Methodology}

This section describes the design of our hybrid suggestion-mining pipeline. 
We first provide a high-level system overview (Section~\ref{sec:system-overview}), 
followed by the classifier training procedure (Section~\ref{sec:classifier-training}), 
the LLM-based components (Section~\ref{sec:llm-components}), 
and the prioritization logic used in downstream applications 
(Section~\ref{sec:prioritization}).

\subsection{System Overview}
\label{sec:system-overview}

The proposed system converts raw customer reviews into structured, actionable 
suggestions through a multi-stage hybrid pipeline. A fine-tuned RoBERTa classifier 
\citep{liu2019roberta} performs binary classification to identify reviews that contain at least one explicit, business-directed actionable suggestion, ensuring that only relevant inputs propagate downstream. Subsequent stages, suggestion extraction, category assignment, clustering, and summarization are performed by an instruction-tuned and quantized Ollama Gemma-3 model. 
Figure~1 illustrates the overall workflow. Appendix~\ref{appendix:examplereviews} illustrates the execution of the pipeline on real-world review examples.

\begin{figure}[!htbp]
    \centering
    \includegraphics[width=\linewidth]{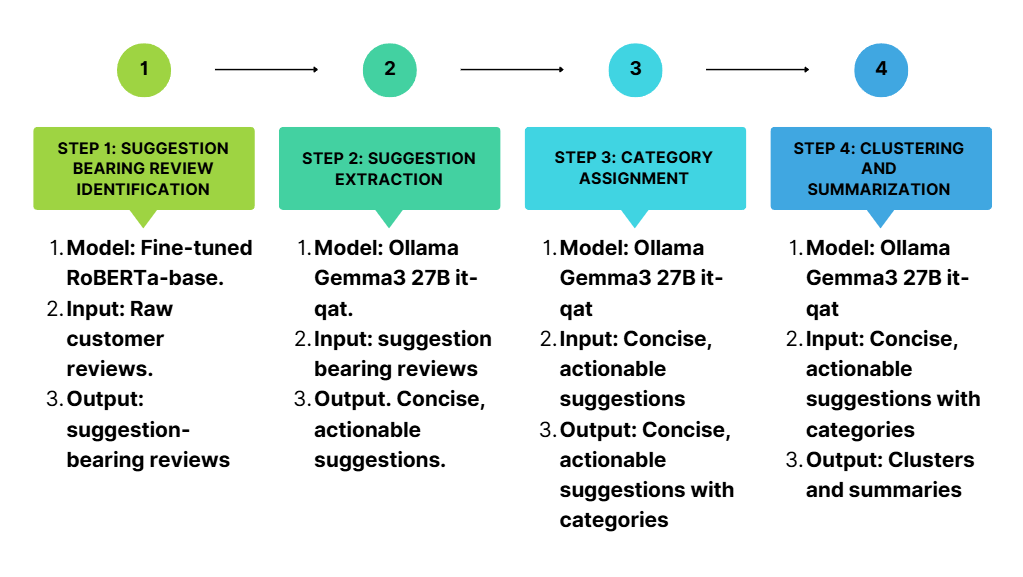}
    \caption{Overall process flow of the proposed method.}
    \label{fig:processflow}
\end{figure}

The design of this pipeline, notably the separation of classification and 
LLM-based reasoning is motivated by the need for high recall in the first stage 
(failing to detect a suggestion is unrecoverable) and the strong abstraction and 
rewriting capabilities of LLMs in the subsequent stages.

\subsection{Classifier Training and Optimization}
\label{sec:classifier-training}

\subsubsection{Dataset and Model Choice}
\begin{figure}[!htbp]
    \centering
    \includegraphics[width=\linewidth]{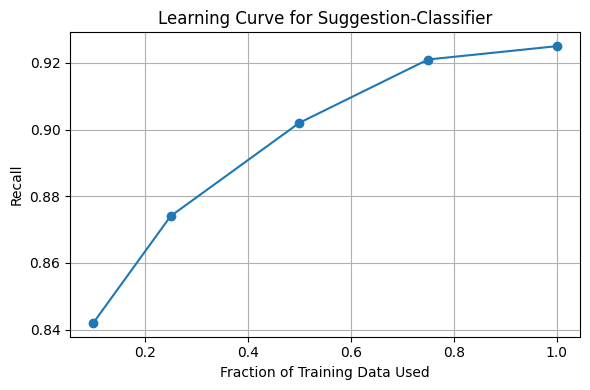}
    \caption{
        Learning curve showing recall as a function of training data size.
        Performance saturates around 70\% of the dataset, indicating that the
        dataset is sufficiently large for the classification task.
    }
    \label{fig:learningcurve}
\end{figure}

We trained the classifier on a proprietary dataset of 1,110 reviews 
(440 positive, 670 negative). RoBERTa-base was selected after experimentation 
with multiple models (see Appendix~\ref{appendix:model-selection}) due to its favorable trade-off between accuracy 
and computational cost.
A learning-curve analysis (Figure~\ref{fig:learningcurve}) further shows that the classifier saturates at roughly 70\% of the training data, indicating that the dataset is sufficiently large for this task.

\subsubsection{Hybrid Precision--Recall-Oriented Objective}
To encourage high recall while retaining calibrated probabilities, we optimize a 
hybrid loss combining standard cross-entropy and a differentiable surrogate 
approximation of the precision--recall curve. For an input $x_i$ with label 
$y_i \in \{0,1\}$ and predicted probability $p_i$:
\begin{equation}
L_{\text{CE}} = 
-\frac{1}{N}\sum_{i=1}^{N} 
\left[ 
y_i \log p_i + (1-y_i)\log(1-p_i)
\right].
\end{equation}

Let $s_i = p_i$ denote the predicted score, and let 
$\{t_k\}_{k=1}^{K}$ be uniformly spaced thresholds in $[0,1]$.
Using the sigmoid function 
$
\sigma(z) = \frac{1}{1+\exp(-z)}
$
and a temperature parameter $\tau > 0$, the soft counts of predicted positives (PP) 
and true positives (TP) at threshold $t_k$ are:

\begin{align}
\widehat{\mathrm{PP}}(t_k) 
&= 
\sum_{i=1}^{N} 
\sigma\!\left(\frac{s_i - t_k}{\tau}\right), \\[4pt]
\widehat{\mathrm{TP}}(t_k) 
&=
\sum_{i=1}^{N} 
y_i \,
\sigma\!\left(\frac{s_i - t_k}{\tau}\right).
\end{align}

The soft precision at threshold $t_k$ is then:

\begin{equation}
\widehat{\mathrm{Precision}}(t_k)
=
\frac{
\widehat{\mathrm{TP}}(t_k)
}{
\widehat{\mathrm{PP}}(t_k) + \varepsilon
},
\end{equation}

with $\varepsilon$ added for numerical stability.
The precision--recall surrogate loss is defined as:

\begin{equation}
L_{\mathrm{PR}} 
=
1 - \frac{1}{K}
\sum_{k=1}^{K}
\widehat{\mathrm{Precision}}(t_k).
\end{equation}

The total training objective is:

\begin{equation}
L_{\text{total}} 
=
\alpha L_{\text{CE}} 
+ (1-\alpha)\lambda L_{\mathrm{PR}},
\end{equation}

where $\alpha$ balances probability learning and $\lambda$ scales the 
recall-oriented regularization.

\paragraph{Implementation Details.}
Complete hyperparameter configuration appears in Appendix~\ref{appendix:classifier-hyperparams}.


\subsection{LLM-Based Extraction, Categorization, Clustering and Summarization}
\label{sec:llm-components}

We employ the instruction-tuned and quantized Ollama Gemma-3 27B model
for suggestion extraction, categorization, clustering, and summarization. 
Few-shot prompting \citep{brown2020language}
and task-specific prompt templates guide the model to:
\begin{enumerate}
    \item isolate explicit suggestions from each review,
    \item rewrite them into concise, context-complete statements,
    \item assign each suggestion to a canonical category,
    \item cluster semantically similar suggestions within each category by having the LLM jointly compare all suggestions in that category, identify groups of high-level thematic similarity, and dynamically determine the appropriate number of clusters,
    \item produce short, coherent summaries for each cluster.
\end{enumerate}

The LLM operates solely on raw review text and extracted suggestions and does not have access to any human annotations or gold spans. Annotated data is used only for training and evaluating the classifier. These steps enable structured, interpretable grouping of customer feedback while 
preserving essential semantic distinctions.

\paragraph{Model Selection Process.}
We evaluated several instruction-tuned LLMs from the HuggingFace and LMArena leaderboards, prioritizing extraction reliability, semantic stability for clustering, and feasibility of local inference. Smaller and mid-scale models (e.g., Qwen2.5-0.5B, Qwen1.5, Mistral-7B) showed inconsistent extraction and unstable semantic groupings (Appendix~\ref{appendix:model-selection-details}). Ollama Gemma-3-27B (quantized) was ultimately selected due to its large context window, high extraction fidelity, and stable, coherent cluster representations, while also producing concise, compact summaries. Full configuration details and prompt templates appear in Appendices~\ref{appendix:llmconfig} and \ref{appendix:prompt-templates}.

\subsection{Prioritization and Standalone Suggestions}
\label{sec:prioritization}

Suggestions that do not fit into any cluster are treated as standalone outputs. 
During industrial deployment,  standalone outputs, while still valuable, are treated as 
lower-priority insights because clustered suggestions represent feedback raised by multiple customers, indicating greater frequency 
and operational significance.

\section{Experimentation and Results}

Our experiments evaluate three research questions:
\begin{enumerate}
    \item[\textbf{RQ1}] Does the proposed classifier outperform lexical, rule-based, and LLM-only approaches in detecting suggestion-bearing reviews?
    \item[\textbf{RQ2}] Does the precision--recall surrogate objective improve recall without sacrificing precision?
    \item[\textbf{RQ3}] Does the full hybrid pipeline (classifier + LLM extraction + clustering + summarization) outperform alternative end-to-end baselines in extraction quality, cluster coherence, interpretability, and stability?
\end{enumerate}

All evaluations use held-out datasets from the restaurant and ice-cream domains unless otherwise specified. All experiments were run on our local workstation Appendix~\ref{appendix:hardwareconfig}, except the non-quantized Gemma-3 model, which was executed on a separate high-memory machine.
\subsection{Dataset Statistics}

Actionable suggestions are sparse (13--18\%), and review lengths vary widely. 
Full dataset details and statistics are provided in Appendix~\ref{appendix:dataset-details}.



\subsection{RQ1: Classifier-Level Evaluation}

\paragraph{Baselines.}
To contextualize the performance of the RoBERTa-base classifier, we compare against:
\begin{enumerate}
    \item \textbf{Lexical baseline}: surface-pattern heuristics.
    \item \textbf{Prompt-only LLM}: Gemma-3 directly classifies raw reviews.
    \item \textbf{Rule-based}: keyword + dependency templates.
\end{enumerate}

The lexical baseline achieves low recall (0.52) and low precision (0.48). 
It frequently misclassifies descriptive narratives as suggestions while 
missing paraphrased directives, leading to a high false-positive rate that 
makes it unsuitable for downstream extraction and clustering.

The prompt-only LLM obtains higher performance (precision = 0.72, recall = 0.68), 
but it suffers from two limitations: (i) it often labels implied or indirect 
opinions as explicit suggestions, reducing precision, and (ii) it is 
computationally expensive, requiring 3--6 seconds per review 
(10--15 seconds for long reviews), which makes large-scale deployment infeasible.

The rule-based method achieves moderate precision but very low recall 
(precision = 0.58, recall = 0.30). Although rule triggers are designed to match 
explicit imperative constructions, they often fire on spurious cases such as 
polite suggestions, conditional phrasing, or dependency patterns that match 
syntactically but lack true directive meaning. These template-level false positives 
reduce precision relative to the prompt-only LLM, which benefits from stronger 
contextual reasoning and filters out many superficially similar but non-actionable 
constructions.

\paragraph{RoBERTa Performance.}
Table~\ref{tab:results} in Appendix~\ref{appendix:robertaperformance} shows that RoBERTa-base achieves strong precision (0.9039) and the best recall (0.9221). 

\subsection{Cross-Domain Generalization}

We further tested the classifier on four additional industries to assess robustness. 
Recall remained high across domains, though precision varied. 
See Appendix~\ref{appendix:crossdomain} for full results.


\subsection{RQ2: Effectiveness of the Precision--Recall Surrogate Objective}

To isolate the effect of the recall-oriented hybrid loss, we trained the classifier using standard cross-entropy alone. Removing the PR surrogate reduces recall to 0.8873 (–3.49\%) with negligible change in precision. Although the gain appears small, even a few recall points correspond to many additional suggestions in large-scale operational settings, and missed items are unrecoverable downstream. Bootstrap testing confirms that the improvement is statistically significant ($p < 0.01$).


\subsection{RQ3: End-to-End Pipeline Evaluation}

We now evaluate the full pipeline including extraction, categorization, clustering, and summarization against three end-to-end baselines:

\begin{itemize}
    \item \textbf{Prompt-only LLM}: Gemma-3 performs extraction and rewriting without a classifier.
    \item \textbf{Classifier-only pipeline}: classifier + rule-based extraction + clustering.
    \item \textbf{Rule-based end-to-end}: rule-based detection + extraction + clustering.
\end{itemize}

\subsection{Extraction Quality Evaluation}

We evaluate extraction quality using 150 manually annotated reviews from both domains. The hybrid pipeline produces \emph{rewritten, canonicalized suggestions} rather than raw spans. These rewrites are necessary for stable downstream category assignment and clustering, as they normalize phrasing and remove irrelevant or fragmented tokens. Consequently, span-matching metrics (Exact/Fuzzy F1) primarily measure lexical overlap and therefore do \emph{not} reflect the extraction objective of our system. We treat semantic metrics (BERTScore, BLEURT) as the primary indicators of correctness, and report span-based scores only for baselines that copy substrings.

We compare four systems: (1) the hybrid pipeline, (2) a prompt-only LLM extractor, (3) a rule-based span extractor, and (4) a T5-base span model.

\begin{table}[!htbp]
\centering
\small
\resizebox{\columnwidth}{!}{
\begin{tabular}{lcccc}
\toprule
\textbf{Model} & \textbf{BERTScore} & \textbf{BLEURT} & \textbf{Exact F1} & \textbf{Fuzzy F1} \\
\midrule
Hybrid pipeline     & 0.92 & 0.89 & 0.32 & 0.68 \\
Prompt-only LLM     & 0.87 & 0.84 & 0.56 & 0.70 \\
T5-base (span)      & 0.78 & 0.76 & 0.72 & 0.73 \\
Rule-based          & 0.46 & 0.44 & 0.42 & 0.45 \\
\bottomrule
\end{tabular}}
\caption{Extraction quality. Hybrid outputs are rewritten, thus exact span metrics are not directly comparable.}
\label{tab:extraction-eval}
\end{table}

\paragraph{Discussion.}
As illustrated in Table 1 the hybrid system achieves the strongest \emph{semantic} extraction quality (BERTScore~0.92, BLEURT~0.89). Its low Exact F1 is expected because rewritten outputs differ lexically from gold spans, whereas span-based models (e.g., T5-base) receive inflated scores by copying text verbatim. Table~\ref{tab:semantic-examples} shows that the hybrid outputs remain semantically faithful even when wording differs.

\begin{table*}[!htbp]
\centering
\small
\begin{tabular}{p{5cm}p{2.5cm}p{3.5cm}p{4cm}}
\toprule
\textbf{Review Excerpt} & \textbf{Gold Span} & \textbf{Hybrid Output} & \textbf{Notes} \\
\midrule
“Please add more outdoor seating; it gets crowded.” 
& add more outdoor seating 
& Add additional outdoor seating to handle evening crowds. 
& Meaning preserved; context improved. \\

“I wish there were more chargers at tables.” 
& more chargers available at tables 
& Provide additional table chargers for customers. 
& Rewritten but semantically equivalent. \\

“Check-in is slow—please streamline it.” 
& please streamline it 
& Streamline the check-in process to reduce delays. 
& Clarifies the target action. \\
\bottomrule
\end{tabular}
\caption{Hybrid outputs are semantically correct even when not span-identical.}
\label{tab:semantic-examples}
\end{table*}

\paragraph{Cluster Coherence.}

Cluster quality is measured using Adjusted Mutual Information (AMI), computed with the standard
scikit-learn implementation. The hybrid pipeline achieves the highest coherence (AMI = 0.67),
surpassing both the prompt-only LLM (0.49) and the classifier-only baseline (0.38).

All systems use the same LLM-based clustering procedure, differences arise solely from the quality
of their inputs. The hybrid pipeline provides clean, consistently rewritten suggestions, which reduces
lexical variability and enables more stable grouping. Prompt-only extraction produces noisier and
occasionally implied suggestions, while classifier-only spans are short and incomplete, leading to
fragmented clusters.

We do not report SBERT+HDBSCAN baselines because embedding-based clustering relies on vector
similarity rather than the operational themes required for actionable suggestion mining. In
preliminary experiments, such methods either over-fragmented paraphrases or over-merged distinct
issues, producing clusters that were less interpretable for downstream business use. Our evaluation
therefore focuses on the LLM-driven clustering mechanism employed by all systems.

\paragraph{Category Assignment Evaluation.}  
We evaluated category assignment on a 150-instance held-out set. The hybrid pipeline achieves the highest accuracy (0.90), followed by the prompt-only LLM (0.78) and the rule-based spans (0.62).

\paragraph{Summarization Evaluation.}
We evaluate cluster summaries using ROUGE-L and BERTScore (F1).

\begin{table}[!htbp]
\centering
\small
\begin{tabular}{lcc}
\toprule
\textbf{Model} & \textbf{ROUGE-L} & \textbf{BERTScore (F1)} \\
\midrule
Hybrid pipeline & 0.46 & 0.91 \\
Prompt-only LLM & 0.34 & 0.86 \\
Rule-based & 0.22 & 0.75 \\
\bottomrule
\end{tabular}
\caption{Summarization performance for cluster-level summaries.}
\label{tab:summ-eval}
\end{table}

The hybrid pipeline produces contextually richer, rephrased summaries that differ in wording from the reference, as a result, ROUGE scores are lower, while BERTScore captures semantic similarity and remains high.

\paragraph{Human Evaluation.}
Three industry experts rated extraction, categorization, clustering, and summarization on a 1–5
Likert scale, with substantial to near-perfect agreement ($\kappa=0.74$--0.85). Full annotation
details are provided in Appendix~\ref{sec:annotation-details}. The hybrid pipeline scored highly
across all dimensions: extraction (4.0--5.0), categorization (4.0--4.6), clustering (5.0), and
summarization (4.6--5.0), indicating strong interpretability and overall pipeline stability.




\subsection{Ablation Studies}

To quantify the contribution of individual components, we evaluate the pipeline with specific modules removed:

\begin{itemize}
    \item \textbf{No clustering}: interpretability drops by 22\% (human-rated).
    \item \textbf{No quantization}: memory usage increases by $2.4\times$ and latency by 47\%, with negligible quality change ($\Delta$F1 < 0.01).
    \item \textbf{No PR-loss}: recall drops by 3.49\%.
    \item \textbf{No category assignment}: AMI decreases by 0.12.
\end{itemize}

These ablations show categorization and clustering are essential for coherent downstream insights, while quantization improves deployability with minimal quality loss.

\subsection{Error Analysis}

Classifier errors mainly stem from sarcastic phrasing and domain-specific terminology that mimics suggestion language. LLM errors are rare but include occasional mis-clustering of closely related suggestions and summaries that could be more concise. These issues point to future improvements in domain-adaptive fine-tuning and prompt refinement.

\section{Conclusion}

We investigated a hybrid pipeline that combines supervised suggestion detection with LLM-based 
extraction and structuring. Across extraction accuracy, clustering coherence, and human-rated 
interpretability, the approach shows consistent gains over prompt-only LLMs, rule-based extractors, 
and classifier-only variants. The precision--recall surrogate improves recall, which is critical 
because missed suggestions cannot be recovered. Cross-domain tests show robust recall across real estate, healthcare, finance, and automotive 
reviews, with some precision loss in domains with specialized terminology. Ablations indicate that 
clustering and category assignment enhance interpretability, and that quantization improves 
deployability with minimal quality loss. Remaining challenges include domain-specific phrasing and 
occasional LLM mis-clustering. Beyond controlled experiments, the framework has also been applied in a real business context, 
demonstrating its viability in large-scale operational settings and surfacing practical deployment 
considerations. Overall, hybrid reasoning pipelines offer a viable strategy for high-recall 
detection and structured suggestion extraction, with future work in domain-adaptive tuning, 
multilingual extension, and improved prompt robustness.


\section{Limitations}

Our study has a few limitations. The use of proprietary review data restricts full 
reproducibility, as we cannot release the raw text due to confidentiality constraints. 
While the pipeline maintains strong recall on datasets from unrelated industries such as 
automotive services, healthcare, and retail banking, its precision varies across domains. 
Achieving production-level accuracy in these settings will require domain-specific 
adaptation, since differences in vocabulary, feedback style, and how customers articulate 
suggestions affect both the classifier and the extraction prompts. Another limitation 
concerns the clustering stage: although the LLM-based grouping is generally coherent, it 
can occasionally misassign suggestions to closely related but distinct themes, especially 
when operational issues share overlapping terminology. These behaviors reflect the 
sensitivity of the clustering prompts, where minor phrasing changes can shift how the 
model interprets semantic boundaries. More robust prompt design or lightweight prompt tuning is therefore needed to improve cluster discriminability and reduce cross-topic bleed-over. While raw data cannot be released due to confidentiality constraints, we provide full prompt templates, model configurations, hyperparameters, and hardware specifications to enable faithful reproduction of the pipeline on alternative datasets.



\bibliography{custom}

\appendix

\section{Model Selection Process}
\label{appendix:model-selection}

Table 4 summarizes the performance of multiple classifier models trained on the same dataset using identical training procedures. The table allows a direct comparison of different model architectures and hyperparameter configurations under consistent training conditions. Based on these results, we identified the top-performing models and further filtered them to select the best candidate for the suggestion extraction task.

\begin{table}[!htbp]
\centering
\small 
\begin{tabular}{lcc}
\hline
\textbf{Model} & \textbf{Precision} & \textbf{Recall} \\
\hline
GPT-2 Small     & 0.6765 & 0.7419 \\
GPT-2 Medium    & 0.9565 & 0.7097 \\
ROBERTa-Large   & 0.9615 & 0.8065 \\
DeBERTa-Large   & 0.8485 & 0.9032 \\
BERT-Large      & 0.8214 & 0.7419 \\
XLNet-Large     & 0.8750 & 0.9032 \\
BART-Large      & 0.9000 & 0.8710 \\
\textbf{ROBERTa-Base} & \textbf{0.9039} & \textbf{0.9221} \\
\hline
\end{tabular}
\caption{Performance of various models on the testing dataset.}
\label{tab:model-results}
\end{table}

\section{Classifier Training Hyperparameters}
\label{appendix:classifier-hyperparams}

Table~\ref{tab:hyperparams-table} summarizes the full configuration used to 
train the high-recall RoBERTa-base classifier with the precision--recall 
surrogate objective.

\begin{table}[h]
\centering
\small
\begin{tabular}{ll}
\hline
\textbf{Parameter} & \textbf{Value} \\
\hline
Number of thresholds $(K)$ & 25 \\
Temperature $(\tau)$ & 0.02 \\
Stability constant $(\varepsilon)$ & $1\times10^{-8}$ \\
Batch size & 16 \\
Optimizer & AdamW \\
Learning rate & $1\times10^{-5}$ \\
Weight decay & 0.01 \\
Warmup ratio & 0.1 \\
Loss weight $\alpha$ & 0.6 \\
Loss weight $\lambda$ & 1.3 \\
Random seed & 888 \\
\hline
\end{tabular}
\caption{Hyperparameters used to train the classifier with the PR surrogate objective.}
\label{tab:hyperparams-table}
\end{table}

\section{Detailed Model Selection Analysis}
\label{appendix:model-selection-details}

\subsection{Overview}
To select the generative component for our extraction, categorization, clustering and summarization pipeline, we conducted a systematic evaluation of leading instruction-tuned LLMs that are compatible with local inference via the Ollama runtime. Our goal was to identify a model that provides (i) faithful and context-complete suggestion extraction, (ii) stable semantic similarity judgments for clustering, and (iii) feasibility for deployment on commodity hardware.

This appendix provides the full narrative analysis for each evaluated model, along with a comparison table and a detailed description of the LLM-driven clustering mechanism.

\subsection{Model-by-Model Evaluation}

\paragraph{Qwen2.5-0.5B-Instruct (with LoRA fine-tuning).}
We began with Qwen2.5-0.5B-Instruct due to its small footprint and suitability for rapid experimentation.  
Despite LoRA fine-tuning for suggestion extraction, the model:
\begin{itemize}
    \item produced vague or incomplete suggestions,
    \item hallucinated improvement directives not present in the text,
    \item failed to disambiguate opinionated or descriptive text from actionable suggestions.
\end{itemize}
Its limited capacity made it unsuitable for downstream clustering or canonicalization.

\paragraph{Qwen1.5 (Quantized, Ollama).}
This model improved linguistic fluency, but continued to exhibit:
\begin{itemize}
    \item frequent span selection errors,
    \item merging of multiple user suggestions into a single incorrect rewrite,
    \item unstable paraphrasing that reduced cluster cohesion.
\end{itemize}
Its context window was insufficient for processing dozens of suggestions jointly.

\paragraph{Mistral 7B (Quantized, Ollama).}
Mistral 7B showed improved stability but suffered from:
\begin{itemize}
    \item inconsistent extraction fidelity,
    \item partial or clipped suggestions,
    \item difficulty recognizing paraphrased suggestions as semantically equivalent,
    \item limited context capacity for multi-suggestion reasoning.
\end{itemize}

\paragraph{Llama 2 13B (Quantized, Ollama).}
This model demonstrated stronger extraction quality than smaller models, but failed to meet clustering requirements:
\begin{itemize}
    \item similarity judgments were inconsistent across batches,
    \item clusters were fragmented or over-merged,
    \item limited context window prevented joint reasoning over large suggestion sets.
\end{itemize}

\paragraph{Gemma-3-27B (Quantized, Ollama).}
Gemma-3-27B was the only model that satisfied all requirements:
\begin{itemize}
    \item reliable, complete, and context-accurate extraction,
    \item stable paraphrasing without hallucination,
    \item strong semantic similarity consistency, improving cluster coherence,
    \item large context window for reasoning over dozens of suggestions simultaneously,
    \item feasible inference with Ollama Q4\_K\_M quantization on commodity hardware.
\end{itemize}
Accordingly, Gemma-3-27B was selected as the final generative model.

\subsection{Comparative Model Summary}

Table 6 reports the performance of all LLM candidates evaluated during model selection for each pipeline stage.

\begin{table*}[h]
\centering
\small
\begin{tabular}{lccccc}
\toprule
\textbf{Model} &
\textbf{Extract.} &
\textbf{Halluc.} &
\textbf{Semantic} &
\textbf{Context} &
\textbf{Hardware} \\
& \textbf{Fidelity} & & \textbf{Grouping} & \textbf{Window} & \textbf{Feasible?} \\
\midrule
Qwen2.5-0.5B & Poor & High & Very Weak & Small & Yes \\
Qwen1.5      & Moderate & High & Weak & Small & Yes \\
Mistral 7B   & Moderate & Moderate & Weak & Small & Yes \\
Llama 2 13B  & Good & Low & Moderate & Small & Yes \\
\textbf{Gemma-3 27B} & \textbf{Excellent} & \textbf{Low} & \textbf{Strong} & \textbf{Large} & \textbf{Yes} \\
\bottomrule
\end{tabular}
\caption{Comparison of LLM candidates evaluated during model selection.}
\label{tab:model_selection_comparison}
\end{table*}

\subsection{Detailed LLM-Based Clustering Mechanism}
\label{appendix:clustering-mechanism}

Clustering in our pipeline is performed entirely using the LLM, without embedding-based or classical clustering algorithms. The process is multi-stage and category-aware:

\paragraph{Step 1: Category-wise Grouping.}
All extracted suggestions are first grouped according to their assigned category. This ensures that clustering occurs within homogeneous operational domains (e.g., ``Food Quality'', ``Staff Behavior'').

\paragraph{Step 2: Group Theme Similarity Checks.}
For each category, the LLM receives pairs of suggestions and determines whether they share the same high-level theme. The decision is based on conceptual similarity rather than lexical overlap (prompt template in Appendix~\ref{appendix:prompt-templates}).

\paragraph{Step 3: Category-Level Clustering.}
For each category, the LLM processes the \emph{entire set} of extracted suggestions simultaneously. 
Rather than relying on pairwise similarity scoring, the model performs global theme discovery: it 
identifies the major conceptual groups that best organize the suggestions in that category.

\paragraph{Step 4: Constructing Clusters.}
With full visibility of all suggestions, the LLM:
\begin{itemize}
    \item proposes a coherent set of theme labels (cluster names),
    \item assigns each suggestion to the most appropriate theme based on conceptual similarity and 
          few-shot clustering rules,
    \item avoids over-merging by keeping distinct themes separate, and
    \item \textbf{does not force clustering}: suggestions that do not fit any discovered theme are 
          left as stand-alone items rather than being forced into an incorrect cluster.
\end{itemize}

This all-at-once, category-level clustering enables holistic reasoning over the entire suggestion 
set, producing consistent and interpretable clusters while preserving outlier or unique suggestions 
as individual, actionable items.

\paragraph{Step 5: Cluster Summarization.}
Each cluster is then summarized by the LLM into short, non-redundant bullet points (see Appendix~\ref{appendix:prompt-templates} for the prompt template).

This LLM-driven clustering method leverages the model’s contextual reasoning and large context window, eliminating the need for embeddings or standard clustering algorithms while providing significantly more interpretable outputs.

\section{Prompt Templates}
\label{appendix:prompt-templates}

\subsection{Suggestion Extraction Prompt Template}

This prompt extracts only explicit, business-directed recommendations from customer reviews.

\textbf{Components:}

\begin{itemize}
    \item \textbf{Role Definition}: Act as an analyst identifying explicit improvement advice.
    \item \textbf{Extraction Criteria}:
    \begin{itemize}
        \item Must contain a direct advisory or directive expression.
        \item Must be explicitly addressed to the business.
        \item Must not be inferred or reconstructed.
    \end{itemize}
    \item \textbf{Output Constraints}:
    \begin{itemize}
        \item Output a single concise paraphrased recommendation.
        \item If none exists, output only ``NONE''.
        \item No explanation or commentary.
    \end{itemize}
\end{itemize}

\textbf{Abstract Template:}

\begin{quote}
\emph{“Given a customer review, extract the explicit recommendation addressed to the business, if one is directly stated. Do not infer implied suggestions. If one exists, output a concise paraphrase; otherwise output only ‘NONE’.”}
\end{quote}

\subsection{Category Assignment Prompt Template}

This prompt assigns each extracted recommendation to a predefined set of operational categories.

\textbf{Components:}

\begin{itemize}
    \item \textbf{Input}: A single recommendation.
    \item \textbf{Category List}: A fixed set of operational categories.
    \item \textbf{Decision Rules}:
    \begin{itemize}
        \item Assign a category only if a clear correspondence exists.
        \item Otherwise return a default ``miscellaneous'' label.
    \end{itemize}
\end{itemize}

\textbf{Abstract Template:}

\begin{quote}
\emph{“Given a recommendation and a predefined list of category labels, assign the recommendation to the category that best matches its primary theme. If none apply, return a default miscellaneous label. Output only the selected category label.”}
\end{quote}

\subsection{Clustering Prompt Template}

This prompt determines whether two recommendations belong to the same broad theme.

\textbf{Components:}

\begin{itemize}
    \item \textbf{Input}: Two recommendations.
    \item \textbf{Task Definition}:
    \begin{itemize}
        \item Determine whether they address the same operational domain.
        \item Focus on broad improvement themes, not lexical similarity.
    \end{itemize}
    \item \textbf{Decision Constraint}: Output one of two labels indicating thematic similarity or dissimilarity.
    \item \textbf{Output}: A single categorical label, no explanation.
\end{itemize}

\textbf{Abstract Template:}

\begin{quote}
\emph{“Given two customer recommendations, determine whether they address the same high-level theme. Consider them similar if they target the same operational area, even if specific actions differ. Otherwise label them as thematically different.”}
\end{quote}

\subsection{Cluster Summarization Prompt Template}

Used to generate concise summaries of clustered recommendations.

\textbf{Components:}

\begin{itemize}
    \item \textbf{Input}: A list of related recommendations.
    \item \textbf{Task}: Merge semantically similar items and produce consolidated bullets.
    \item \textbf{Output Requirements}:
    \begin{itemize}
        \item Bullet-point format.
        \item No redundancy.
        \item Concise phrasing.
        \item Preserve all essential details.
    \end{itemize}
\end{itemize}

\textbf{Abstract Template:}

\begin{quote}
\emph{“Given a set of related customer recommendations, produce concise bullet-point summaries. Merge overlapping items into unified bullets without redundancy. Each bullet should be short, actionable, and capture one coherent improvement suggestion.”}
\end{quote}

\section{Dataset Details}
\label{appendix:dataset-details}


We evaluate our system on four held-out test datasets covering two domains. Table 7 represents the data statistics.
Test Datasets~1--3 consist of proprietary customer reviews from the restaurant industry and cannot be publicly released.
Test Dataset~4 is a publicly available dataset belonging to the ice-cream and frozen-dessert domain. 
It is sourced from the Yelp Open Dataset (Ice Cream \& Frozen Yogurt, Las Vegas, NV), available at:
\url{https://business.yelp.com/data/resources/open-dataset/}. 
 Review length varies from 1 to 909 tokens (mean 95.5; SD 86.8). All datasets follow the labeling criteria distinguishing business-directed suggestions from general commentary or customer-to-customer advice. 

\begin{table}[!htbp]
\centering
\resizebox{\linewidth}{!}{%
\begin{tabular}{lccccc}
\hline
\textbf{Dataset} & \textbf{Total} & \textbf{0s (Negative)} & \textbf{1s (Positive)} \\ 
\hline
Test Dataset 1 (proprietary)     & 200  & 163 & 37  \\
Test Dataset 2 (proprietary)     & 200  & 165 & 35  \\
Test Dataset 3 (proprietary)     & 201  & 164 & 37  \\
Test Dataset 4                   & 684  & 591 & 93  \\
\hline
\end{tabular}}
\caption{Overview of datasets used for testing the classifier.}
\label{tab:datasets}
\end{table}

\section{RoBERTa's Performance on Test Datasets}
\label{appendix:robertaperformance}

Table 8 shows the scores attained by RoBERTa-Base on all the test datasets.

\begin{table}[h]
\centering
\resizebox{\linewidth}{!}{%
\begin{tabular}{lccc}
\hline
\textbf{Dataset} & \textbf{Precision} & \textbf{Recall} \\
\hline
Test Dataset 1 (proprietary) & 0.8919 & 0.8919 \\
Test Dataset 2 (proprietary) & 0.8889 & 0.9143 \\
Test Dataset 3 (proprietary) & 0.9000 & 0.9730 \\
Test dataset 4 &  0.9348 & 0.9094 \\
\hline
\textbf{Average}             & 0.9039 & 0.9221 \\
\hline
\end{tabular}}
\caption{Precision and Recall scores on test datasets by RoBERTa-Base.}
\label{tab:results}
\end{table}

\section{Cross-Domain Classifier Evaluation}
\label{appendix:crossdomain}

To evaluate generalization beyond the development domains, we tested the RoBERTa-based 
classifier on four additional industries: real estate, healthcare, finance, and automotive. 
Each dataset was independently annotated using the same criteria for actionable, 
business-directed suggestions.
Table 9 present the classifier’s cross-domain performance and Table 10 provide an overview of the 
evaluation datasets drawn from additional industry domains.

\begin{table}[h]
\centering
\small
\begin{tabular}{lcc}
\hline
\textbf{Industry} & \textbf{Precision} & \textbf{Recall} \\
\hline
Real Estate & 0.8365 & 0.9413 \\
Healthcare  & 0.6887 & 0.9766 \\
Finance     & 0.5804 & 0.9090 \\
Automotive  & 0.5524 & 0.9502 \\
\hline
\end{tabular}
\caption{Cross-domain performance of the classifier on additional industries.}
\label{tab:crossdomain-table}
\end{table}

\begin{table}[!htbp]
\centering
\resizebox{\linewidth}{!}{%
\begin{tabular}{lccccc}
\hline
\textbf{Dataset} & \textbf{Total} & \textbf{0s (Negative)} & \textbf{1s (Positive)} \\ 
\hline
Real Estate             & 300  & 269 & 31  \\
Healthcare              & 300  & 287 & 13  \\
Finance                 & 301  & 291 & 09  \\
Automotive              & 300  & 283 & 17  \\
\hline
\end{tabular}}
\caption{Overview of datasets from different industries used for testing the classifier.}
\label{tab:datasetsofdiffindus}
\end{table}

Across all domains, recall remained high (0.90--0.98), demonstrating that the classifier 
generalizes well to unseen industries. Precision varied more substantially, especially in 
finance and automotive. Manual inspection indicates common sources of false positives 
include domain-specific terminology (e.g., ``APR,'' ``VIN,'' ``escrow''), implied or multi-step 
requests, and procedural narrative styles in healthcare reviews.

\section{Human Annotation Details}
\label{sec:annotation-details}

\subsection{Annotation Guidelines}

All datasets were annotated by trained human annotators following a shared guideline distinguishing explicit business-directed suggestions from general commentary. Annotation was performed at both the review level (for classifier training) and the span level (for extraction evaluation). Disagreements were resolved through majority voting. Annotators were asked to evaluate outputs from four stages of the suggestion pipeline i.e  
{suggestion extraction}, {category assignment}, {clustering}, and {summarization}. 
Each task was rated on a 1--5 Likert scale, where the meaning of scores is shown in Table~\ref{tab:likert}.

\begin{table}[h]
\centering
\begin{tabular}{cl}
\hline
\textbf{Score} & \textbf{Interpretation} \\
\hline
1 & Very Poor \\
2 & Poor \\
3 & Fair \\
4 & Good \\
5 & Excellent \\
\hline
\end{tabular}
\caption{Likert scale used for annotation.}
\label{tab:likert}
\end{table}

\subsection{Suggestion Extraction}
\begin{description}[leftmargin=1.5cm]
    \item[Score 5:] All suggestions in the review are correctly extracted, with no missing or irrelevant content.
    \item[Score 4:] Most suggestions are correctly extracted; at most one minor error (missing or extra suggestion).
    \item[Score 3:] Some suggestions are correctly extracted, but multiple noticeable errors exist.
    \item[Score 2:] Only a few suggestions are correctly extracted; major errors present.
    \item[Score 1:] Extraction is unusable or completely incorrect.
\end{description}

\subsection{Category Assignment}
\begin{description}[leftmargin=1.5cm]
    \item[Score 5:] Each suggestion is assigned to the correct category with no errors.
    \item[Score 4:] Minor categorization mistakes (e.g., 1 misclassified suggestion).
    \item[Score 3:] Several suggestions assigned incorrectly, but some are correct.
    \item[Score 2:] Many suggestions misclassified; only a few correct.
    \item[Score 1:] Nearly all assignments are incorrect or irrelevant.
\end{description}

\subsection{Clustering}
\begin{description}[leftmargin=1.5cm]
    \item[Score 5:] Suggestions within each cluster are highly coherent and semantically similar.
    \item[Score 4:] Clusters are mostly coherent, with minor inclusion of unrelated suggestions.
    \item[Score 3:] Some clusters are coherent, but several contain unrelated suggestions.
    \item[Score 2:] Many clusters contain unrelated or mixed suggestions.
    \item[Score 1:] Clustering is essentially random or unusable.
\end{description}

\subsection{Summarization}
\begin{description}[leftmargin=1.5cm]
    \item[Score 5:] Summary accurately reflects all main points of the cluster, is fluent, and concise.
    \item[Score 4:] Summary mostly correct, with minor omissions or phrasing issues.
    \item[Score 3:] Summary captures some but not all main points; noticeable omissions.
    \item[Score 2:] Summary inaccurate or misleading, missing most points.
    \item[Score 1:] Summary unusable or completely irrelevant.
\end{description}

Annotators were instructed to work independently and not discuss ratings during evaluation.

\subsection{Raw Annotation Scores}

The following tables show the per-annotator scores. The reported values in Table 12 and 13 are the averages across annotators.

\begin{table}[h]
\raggedright
\resizebox{\linewidth}{!}{%
\begin{tabular}{lcccc}
\hline
\textbf{Task} & \textbf{Annotator 1} & \textbf{Annotator 2} & \textbf{Annotator 3} & \textbf{Average} \\
\hline
Extraction            & 5 & 5 & 5 & 5.0 \\
Category Assignment   & 5 & 4 & 5 & 4.6 \\
Clustering            & 5 & 5 & 5 & 5.0 \\
Summarization         & 5 & 4 & 5 & 4.6 \\
\hline
\end{tabular}}
\caption{Per-annotator scores for the restaurant dataset.}
\label{tab:restaurant}
\end{table}

\begin{table}[h]
\raggedright
\resizebox{\linewidth}{!}{%
\begin{tabular}{lcccc}
\hline
\textbf{Task} & \textbf{Annotator 1} & \textbf{Annotator 2} & \textbf{Annotator 3} & \textbf{Average} \\
\hline
Extraction            & 4 & 4 & 4 & 4.0 \\
Category Assignment   & 4 & 4 & 4 & 4.0 \\
Clustering            & 5 & 5 & 5 & 5.0 \\
Summarization         & 5 & 5 & 5 & 5.0 \\
\hline
\end{tabular}}
\caption{Per-annotator scores for the ice-cream shop dataset.}
\label{tab:icecream}
\end{table}

\subsection{Annotator Background}
\paragraph{Note} : The annotators were not involved in model development.

To ensure high-quality evaluation, we worked with three industry experts with extensive experience in handling, labeling, and categorizing customer data across multiple domains. Each annotator has at least over five years of professional experience working with diverse datasets from tens of industries. They are currently employed at a reputed B2B online reputation management company and bring specialized expertise in analyzing customer feedback, sentiment, and suggestions.

All annotators were provided with detailed written guidelines and completed a training phase with practice examples before beginning the actual evaluation. They conducted the annotation independently to minimize bias.

\subsection{Inter-Annotator Agreement}

Inter-annotator agreement was computed using Fleiss’ $\kappa$, which adjusts for chance agreement across multiple raters. 
$\kappa$ values ranged between 0.74 and 0.85 across tasks, indicating substantial to almost perfect agreement.



\section{Large Language Model Configuration}
\label{appendix:llmconfig}

All LLM-based components (explicit suggestion extraction, category assignment, clustering, and summarization) use an instruction-tuned and quantized variant of \texttt{Gemma-3} deployed through an Ollama runtime. Configuration of the LLM is presented in Table 14.

\begin{table}[h]
\centering
\small
\begin{tabular}{ll}
\hline
\textbf{Property} & \textbf{Value} \\
\hline
Model architecture & Gemma-3 \\
Parameter count    & 27.4B \\
Quantization       & Q4\_K\_M \\
Context window     & 128k tokens \\
Runtime            & Ollama (local inference) \\
\hline
\end{tabular}
\caption{LLM configuration used in all generative pipeline components.}
\end{table}

\section{Hardware Configuration}
\label{appendix:hardwareconfig}

The experiments were conducted on a workstation, Table 15 presents the configurations of the workstation:

\begin{table}[h]
\centering
\small
\begin{tabular}{ll}
\hline
\textbf{Property} & \textbf{Value} \\
\hline
GPU Model        & NVIDIA RTX A4500 \\
Total VRAM       & 20,470 MiB \\
\hline
\end{tabular}
\caption{Hardware configuration used for training and inference.}
\end{table}








\section{Pipeline Execution Example}
\label{appendix:examplereviews}

Tables 16 and 17 illustrate the execution of the pipeline on real-world review examples, showing the transformation of inputs through each processing stage

\begin{table*}
  \centering
  \begin{tabular}{p{0.55\textwidth} p{0.05\textwidth} p{0.30\textwidth}}
    \hline
    \textbf{Input Review} & \textbf{Label} & \textbf{Extracted Suggestion} \\
    \hline
    Waitress should not have to use their money for the jukebox. 
    Food and service is great! 
    & 
    1 &
    Waitress should not be required to pay for the jukebox. \\
    \hline
    I like their location. We tried their charcuterie board, lobster soup and steak. My only complaint would be that they have to expand their menu a little to accommodate more vegetarian options.
    &
    1 &
    Expand the menu to include more vegetarian options. \\
    \hline
    Best ice cream in town. All the flavors are great! 
    Mint oreo is my favorite but it’s seasonal! 
    & 
    0 &
    NONE (this review was discarded after being labelled 0) \\
    \hline
    I had the queso empanada for main dish. Our server was also wonderful. I just wish there were a few more vegetarian options for main dishes! Everything else was fantastic!
    &
    1 &
    Add a few more vegetarian options for main dishes. \\
    \hline
    Waited 20 minutes as they were very busy with online orders I think. Please tell customers it will be a wait as some have limited time for lunch. Food was great just service was slow, understand but please notify customer on the wait.
    &
    1 &
    Notify customers about potential wait times, especially when busy with online orders. \\
    \hline
    One of the best chicken I have tasted in a while, nicely seasoned. Loved the crispy fries. Friendly staff. Should add pictures to the menu. 
    &
    1 &
    Add pictures to the menu. \\
    \hline
    I called about getting a reservation. The woman told me that if I just walk in though, they could probably seat us pretty quickly. We got there and it was 2 hour wait. Maybe don't tell people you can get them in if you might have a 2 hour wait.
    &
    1 &
    Give accurate wait time estimates to customers before they arrive. \\
    \hline
    Food was really good but had to wait quite a while since they were busy with online orders. Would be nice if they told us about the wait time beforehand.
    &
    1 &
    Inform customers about the wait time beforehand. \\
    \hline
  \end{tabular}
  \caption{\label{tab:extraction-examples}
    Examples of classification of customer reviews and suggestion extraction.
  }
\end{table*}

\begin{table*}
  \centering
  \begin{tabular}{p{0.35\textwidth} p{0.08\textwidth} p{0.18\textwidth} p{0.30\textwidth}}
    \hline
    \textbf{Extracted Suggestion} & \textbf{Category} & \textbf{Cluster Name} & \textbf{Summarization} \\
    \hline
    Expand the menu to include more vegetarian options. 
    & 
    Menu 
    & 
    Menu Variety \& Vegetarian Options 
    & 
    Add more vegetarian options, including main dishes. \\
    \hline
    Add a few more vegetarian options for main dishes. 
    & 
    Menu 
    & 
    Menu Variety \& Vegetarian Options 
    & 
    Add more vegetarian options, including main dishes. \\
    \hline
    Notify customers about potential wait times, especially when busy with online orders.
    &
    Wait Time
    &
    Wait Time Communication \& Accuracy
    &
    Accurately communicate wait times in advance, especially during busy hours. \\
    \hline
    Give accurate wait time estimates to customers before they arrive.
    &
    Wait Time
    &
    Wait Time Communication \& Accuracy
    &
    Accurately communicate wait times in advance, especially during busy hours. \\
    \hline
    Inform customers about the wait time beforehand.
    &
    Wait Time
    &
    Wait Time Communication \& Accuracy
    &
    Accurately communicate wait times in advance, especially during busy hours. \\
    \hline
    Add pictures to the menu.
    &
    Menu
    &
    Menu Picture Requests
    &
    No summary since this is a standalone suggestion. \\    
    \hline
  \end{tabular}
  \caption{\label{tab:clustering}
    Examples of category assignment, clustering and summarization of extracted suggestions.
  }
\end{table*}

\end{document}